*Research Article*

# Real-Time Human Detection for Aerial Captured Video Sequences via Deep Models

**Nouar AlDahoul, Aznul Qalid Md Sabri, and Ali Mohammed Mansoor**

*Faculty of Computer Science and Information Technology, University of Malaya, Kuala Lumpur, Malaysia*

Correspondence should be addressed to Nouar AlDahoul; nouar.aldahoul@live.iium.edu.my
and Aznul Qalid Md Sabri; aznulqalid@um.edu.my

Human detection in videos plays an important role in various real life applications. Most of traditional approaches depend on utilizing handcrafted features which are problem-dependent and optimal for specific tasks. Moreover, they are highly susceptible to dynamical events such as illumination changes, camera jitter, and variations in object sizes. On the other hand, the proposed feature learning approaches are cheaper and easier because highly abstract and discriminative features can be produced automatically without the need of expert knowledge. In this paper, we utilize automatic feature learning methods which combine optical flow and three different deep models (i.e., supervised convolutional neural network (S-CNN), pretrained CNN feature extractor, and hierarchical extreme learning machine) for human detection in videos captured using a nonstatic camera on an aerial platform with varying altitudes. The models are trained and tested on the publicly available and highly challenging UCF-ARG aerial dataset. The comparison between these models in terms of training, testing accuracy, and learning speed is analyzed. The performance evaluation considers five human actions (digging, waving, throwing, walking, and running). Experimental results demonstrated that the proposed methods are successful for human detection task. Pretrained CNN produces an average accuracy of 98.09%. S-CNN produces an average accuracy of 95.6% with soft-max and 91.7% with Support Vector Machines (SVM). H-ELM has an average accuracy of 95.9%. Using a normal Central Processing Unit (CPU), H-ELM's training time takes 445 seconds. Learning in S-CNN takes 770 seconds with a high performance Graphical Processing Unit (GPU).

## 1. Introduction

Human detection in videos (i.e., series of images) plays an important role in various real life applications (e.g., visual surveillance and automated driver assistance). The task of human detection in a series of aerial images is challenging due to various reasons. One of these reasons is the variation of human size in the video frame. This results from changing the altitude of the platform that the camera is attached to during the task. Accuracy and short training time are the two important factors that should be taken into consideration to get a robust human, nonhuman classification system.

The process of feature design and selection is so sensitive because it has a significant role in improving the performance of the model (the classifier in this work). Handcrafted methods are specific domain knowledge. Some features that are good for one application are not discriminative for other applications. Besides that, these features need to be designed by an expert under the process of feature engineering. On the other hand, feature learning methods are fruitful when the size of datasets is large with thousands of classes because they are domain adaptation. They are able to learn highly abstract features automatically without human intervention and directly from raw pixels without designing specific features. This advantage is very fruitful when there is a lack of an expert's knowledge. On top of that, they are robust against dynamical events in different scenarios: the convolutional neural network (CNN), which is one of the supervised feature learning methods, extracts spatial structure by using convolutions that provide local representations, pooling that is shift-invariant, and normalization that is adapted to illumination change. Hierarchical extreme learning machine (H-ELM), which is one of the unsupervised feature learning methods, utilizes sparse autoencoders to provide more robust features that adapt with data variations without preprocessing. Deep models have proven to be proficient in human and nonhuman classification.

Several papers have already utilized handcrafted features for human detection and have demonstrated that these features are useful and successful for specific tasks. Local binary patterns (LBP) [1] or Scale-Invariant Feature Transform (SIFT) [2] were used for feature extraction in the given image. Using these features, different classifiers were trained. Examples are random forest and support vector machine. Histogram of gradient (HOG) is also a prominent example [3]. HOG preserves information concerning the gradient surrounding detected spatial interest points [3]. HOG was proven to be robust against changes in human appearance [3]. SVM was trained utilizing HOG features. Fastest Pedestrian Detector of the West (FPDW) was also one of the methods that depended on engineered features [4]. It utilizes HOG at different scales. Its speed outperforms state of the art methods by one order of magnitude but it preserves accuracy. The primary strength of FPDW is that it only requires a single frame for human detection but it requires a large number of pixels.

There are two main types of feature learning approaches: supervised and unsupervised. A supervised learner needs input/output pairs to learn the features. An unsupervised learner uses only inputs to find its features. S-CNN is a supervised feature learner that fine-tunes the model parameters iteratively. AlexNet is a supervised CNN model that has already been trained on huge datasets and can be used as a feature extractor. HELM is an unsupervised learner that generates input weights randomly and calculates output weights analytically. These three different models were applied in this paper for the purpose of human, nonhuman classification.

CNNs are multilayer perceptron neural networks. They contain a number of convolutional layers, pooling layers, fully connected layers, and normalization layers. They were inspired by the biological process of organizing the animal visual cortex [5]. The weights of the whole layers in the networks are fine-tuned to produce specific classes. The network is trained by using the Stochastic Gradient Descent (SGD) algorithm [5]. CNN has been employed in different applications such as face recognition [6]. In this system, various challenging factors such as pose change, illumination variation, and partial occlusion exist. In [6], a genetic algorithm was used to find the optimal CNN structure. An ensemble of SVM was used for classification. CNN has also been utilized to solve various problems in natural language processing such as sentence modelling, classification, and prediction [7–9]. Three-dimensional CNN was used in the medical field to segment the brain tumor in MR images [10]. An extension of CNN to 3D was utilized for action recognition in [11]. Spatiotemporal features were automatically learned. A Recurrent Neural Network (Long Short-Term Memory) was trained for sequential classification. In this paper, supervised CNN is demonstrated to learn discriminative features that are being applied to the classifiers.

Different CNN models have already been trained with big datasets. One of them is AlexNet. The target dataset was ImageNet. AlexNet was used to classify 1.2 million high resolution images to 1000 different classes [12]. This network achieved a very high accuracy. A big neural network with 60 million connections and 650,000 nodes was built. It includes five convolutional layers, max pooling layers, three fully connected layers, and a 1000-class soft-max layer. The dropout regularization method was utilized to reduce overfitting in the fully connected layers. This architecture was utilized as a feature extractor in different studies. The last fully connected layer was removed. Various classifiers were connected directly to the output of the seventh layer for classification purposes. AlexNet model is used in this paper to extract the features before applying them on the classifiers.

Hierarchical extreme learning machine is an efficient fast deep model that was used to learn features automatically [13]. HELM has been involved in different applications such as digit classification, car detection, tracking, and gesture recognition. It was found to outperform the state of the art in terms of training speed by one order of magnitude. It was able to increase the speed of learning because there is no need to fine-tune the weights iteratively. In this model, the biases and input weights are generated randomly, but the output weights are calculated analytically. HELM model is also utilized in this paper to learn the features in an unsupervised way before applying them on the classifiers.

Various applications have an embedded human detection concept. One of them was the driving assistance systems [14]. A camera was used instead of Light Detection and Ranging (LIDAR) to detect objects in a single frame. The LIDAR sensor bounces laser beams to determine the distance. It is expensive, sensitive to temperature, and good for only short distances. The camera is the cheaper replacement that is not affected by distance and temperature. In this system, automatic feature learning via fast deep network cascades was used to perform human detection. It was tested on the Caltech dataset in videos captured by a camera mounted in a street. CNN was also utilized in driving assistance system to detect humans [15]. The detection system was installed in a vehicle. In this situation, a low cost and high accuracy technology is required. CNN was merged with random dropout and ensemble inference network (EIN) to improve the generalization performance [15]. In this paper, the human detection task is demonstrated as a challenging task. The image samples vary in activities, positions, orientations, viewpoints, cloths color, and scale. The altitude of the camera is also varied according to the moving airborne platform.

Using an aerial platform to perform human detection has been in the attention of researchers for a significant period of time. Feature engineering methods were used for this objective. A framework that is based on optical flow and graph representation was employed to extract the moving areas from the frames of moving cameras in the Predator Unmanned Airborne Vehicle (UAV) [16]. A dynamic template was used to merge connected graph components to describe the graph completely. Another human detection approach is based on appearance [17]. It was proposed to detect humans from a high altitude in an aerial image. An enhanced version of the Haar features was utilized to characterize the object shape. For color data, rectangular features were used. Small persons (including their shadows) were detected utilizing an AdaBoost binary classifier. The combination of FPDW and Moving Object Detection (MOD) was utilized for UCF-ARG aerial dataset in [18]. MOD is based on the idea of detecting all moving objects in the moving background. The quality of stabilization in this situation is important to detect moving clusters of pixels. In this work, feature learning methods are utilized for human detection using the same UCF-ARG aerial dataset.

Optical flow model was used for human detection in various applications. The task of mobile robot navigation utilized optical flow to detect humans in real time when the robot is moving [19]. The thermal infrared camera was used to achieve this goal. Human motion detection using a combination of optical flow and classification methods was proposed in [20]. Region segmentation was used in the first stage and after getting geometric flow, feature extraction was employed by Bandelet transform. Supervised learning was added in the last stage for classification. In this paper, optical flow is utilized as a first stage to detect moving objects with moving camera. The patches of moving objects are stored as training samples including human and nonhuman patches. These samples are used to learn the features and then to train the classifiers.

Several papers have already utilized handcrafted features for human detection. This paper does not focus on these methods and does not try to compare the handcrafted methods with deep model based methods. The objective of this paper is to study and compare different deep learning methods to detect humans in a challenging scenario that includes a camera attached to a moving airborne object. The comparison between three different deep models which are supervised CNN, pretrained CNN, and HELM is demonstrated for feature learning and model building for the UCF-ARG aerial dataset. An optical flow model is added as a first stage in the three systems to get the training and testing samples as inputs to deep models.

The novelty of our work is as follows:

(i) To the best of our knowledge, this work is the first one that utilizes different deep models for the public UCF-ARG aerial dataset for human detection.

(ii) Supervised CNN is demonstrated to find optimal features that are discriminative to two classes of human and nonhuman. Soft-max and SVM are used in the last layer of CNN to produce the classification output.

(iii) Pretrained AlexNet CNN model that has already been trained on ImageNet dataset for visual object recognition to classify 1000 different classes is demonstrated as a feature extractor with fixed parameters after removing the fully connected layers to find discriminative features for human, nonhuman classification.

(iv) HELM is also discussed to take into consideration the trade-off between high accuracy and low training time.

(v) The comparison between CNN as a supervised feature learner, pretrained CNN as a feature extractor, and HELM as an unsupervised feature learner in terms of learning speed and accuracy is evaluated for five human actions (digging, waving, throwing, walking, and running).

The organization of the paper is as follows: In Section 2, the three proposed systems that consist of the optical flow model and three deep models are described. Section 3 discusses the experimental results and analyzes them in terms of training speed and accuracy. Section 4 demonstrates the efficiency of the proposed system by summarizing the outcome of this work.

## 2. The Methodology

Visual aerial data is first captured by a moving camera mounted on an airborne platform. In this work, we work with the publicly available and challenging UCF-ARG aerial dataset [21]. The frames of the videos are used as input for the optical flow stabilization approach. Image patches of objects resulting from the computation of the optical flow will be produced for both human and nonhuman objects. Nonhuman patches include cars, grass, tools, and other different regions from the background.

The three deep models studied in our work (i.e., supervised CNN, pretrained CNN, and HELM) will then utilize these patches as inputs. The output of both deep models is an efficient type of representation for the objects. These representations are then classified into binary classes (human and nonhuman) using soft-max or support vector machine (SVM) in supervised CNN and pretrained CNN and extreme learning machine (ELM) in HELM. A brief review about each module used in the proposed detection system (optical flow, supervised CNN, pretrained CNN, ELM, and HELM) is summarized in the following subsections. A block diagram of the proposed systems is shown in Figure 1.

*2.1. Background Stabilization by Optical Flow Model [22, 23].* The first stage in our proposed system, which is background stabilization, is done using the optical flow model. Optical flow estimates the speed and direction of the motion vector between sequences of frames. This stage is important because it tackles the camera movement issue that results from the moving aerial platform. Feature images are produced by thresh-holding and performing a morphological operation (closing) to the motion vectors. Blob analysis is then performed to locate moving objects in each binary feature image. Next, green boundary boxes are overlaid surrounding the detected objects. The quality of optical flow for background stabilization is important as it is the first stage, before feature learning is performed via deep models which act as input for the classifiers.

To find the optical flow between two frames, two optical flow constraint equations are used:

$$I_t + I_x h + I_y v = 0, \quad (1)$$

where $I_t$, $I_x$, and $I_y$ are the derivatives of spatiotemporal brightness for a frame, $v$ is the vertical part of optical flow, and $h$ is the horizontal part.

When optical flow is applied over the whole frame, the Horn-Schunck approach [23] finds the velocity field

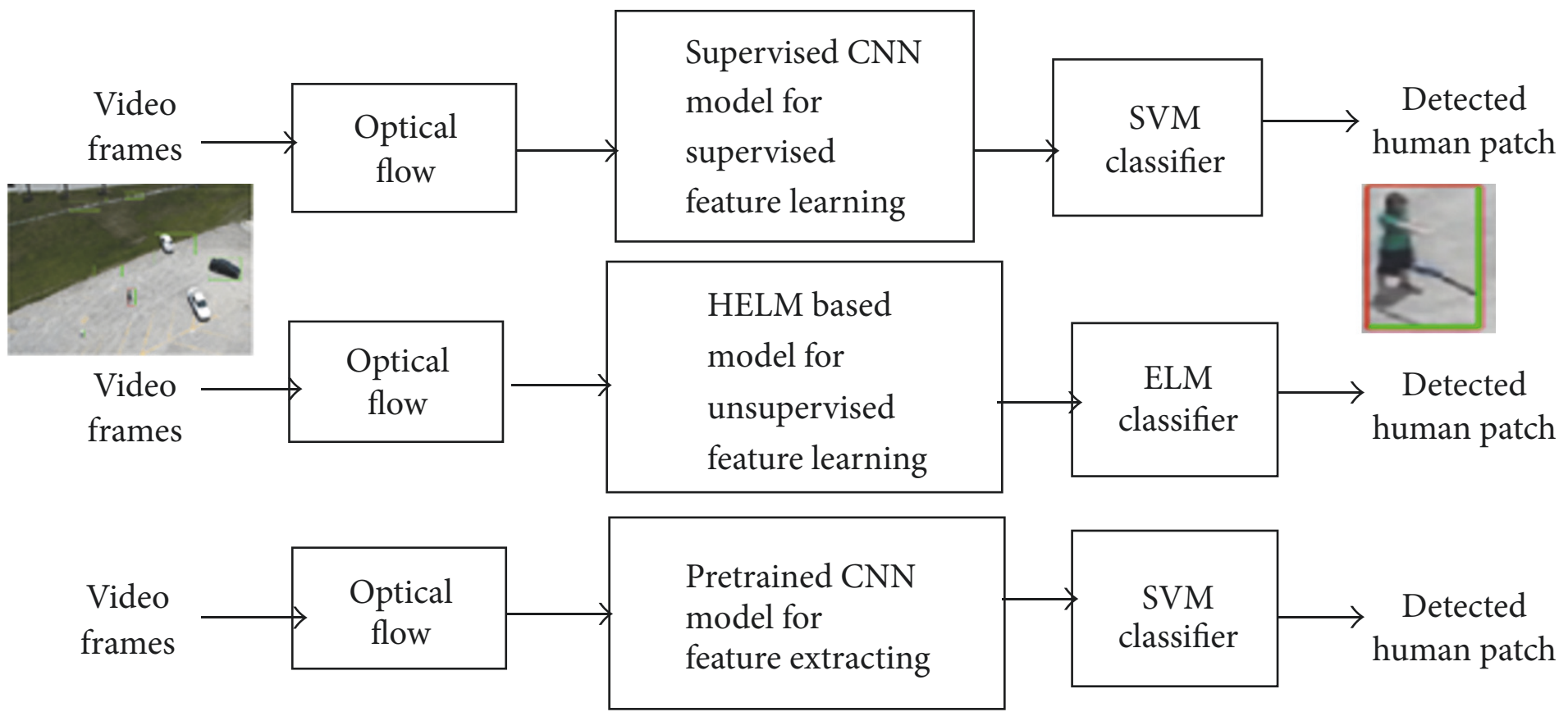

FIGURE 1: Block diagram of the three proposed systems for human detection with aerial data.

estimation for each pixel in the frame, $[h \quad v]^T$, that minimizes the following equation:

$$E = \iint \left(I_t + I_x h + I_y v\right)^2 dx\, dy + \alpha \iint \left\{ \left(\frac{dh}{dx}\right)^2 + \left(\frac{dh}{dy}\right)^2 + \left(\frac{dv}{dx}\right)^2 + \left(\frac{dv}{dy}\right)^2 \right\}, \quad (2)$$

where $\alpha$ is a scaling factor for the optical flow computation and $dh/dx$ and $dh/dy$ are the spatial derivatives of the optical velocity $h$

$$h^{k+1}_{x,y} = \overline{h}^{k}_{x,y} - \frac{I_x \left[I_x \overline{h}^{k}_{x,y} + I_y \overline{v}^{k}_{x,y} + I_t\right]}{\alpha^2 + {I_x}^2 + {I_y}^2},$$
$$v^{k+1}_{x,y} = \overline{v}^{k}_{x,y} - \frac{I_y \left[I_x \overline{h}^{k}_{x,y} + I_y \overline{v}^{k}_{x,y} + I_t\right]}{\alpha^2 + {I_x}^2 + {I_y}^2}. \quad (3)$$

In these equations, $[h^k_{x,y} \quad v^k_{x,y}]$ is the velocity estimate at pixel $(x, y)$ and $[\overline{h}^k_{x,y} \quad \overline{v}^k_{x,y}]$ is the neighborhood average of $[h^k_{x,y} \quad v^k_{x,y}]$. The initial velocity is 0 when $k = 0$. The Horn-Schunck method is used to solve $h$ and $v$.

*2.2. Deep Model Based Human Detection.* After applying optical flow to stabilize the scenes, humans are detected utilizing a deep model approach for human and nonhuman classification purposes.

*2.2.1. Convolutional Neural Network [5, 6].* The Convolutional Neural Network (CNN) was inspired by biological evidences in the visual cortex of mammal brains. CNN is a supervised neural network that has multiple layers. It is able to extract the efficient features during the training stage. Thus, in this work, it functions optimally in learning important and discriminative features from a series of images (i.e., videos) containing humans. Its generalization capability enables it to work well even for an arbitrary series of images. CNN is invariant to shift, rotation, and scale. This is due to the pooling layers in CNN which reduce the number of locally or specially represented features and control the overfitting problem. CNN is robust to misrepresented input data because the structure of deep layers produces a highly abstract representation at the final layer. It is also robust to misbalanced data that results from the unequal division of data into different classes. The local receptive fields, spatial subsampling, and weight sharing are the main properties that make this method so robust. The weight sharing property can reduce the number of parameters and boost the generalization performance.

The CNN structure includes two main blocks: automatic feature learning and a classifier. The CNN works by putting a raw image as input. Features are learned by using multiple layers of feature maps. The convolutional filters are applied on raw images. Next, down sampling is performed to reduce the size in each layer. Different structures of CNN exist. They often differ from each other in the number of feature maps in every convolutional layer, the dimensions of convolution filters, the specific connection between layers, and the activation functions.

*(1) Supervised CNN Model [24, 25].* In the supervised CNN model, there are many fully connected layers that are connected before the classification layer. The weights of the whole layers in the networks are fine-tuned to produce specific classes related to the task. The learned features are discriminative to these classes. This is called a supervised feature learning approach.

The S-CNN network was trained by using the Stochastic Gradient Descent (SGD) algorithm with momentum [25]. In this algorithm, parameters (weights and biases) are updated in one step to minimize the error function (loss). Small steps are added to negative gradient.

$$Q\theta_{k+1} = \theta_k - \alpha \nabla E\left(\theta_k\right), \quad (4)$$

where $k$ is the number of iterations, $\alpha$ is the learning rate, $\theta$ is the vector of parameters, $E(\theta)$ is the loss function, and $\nabla E(\theta)$ is the gradient.

The entire training dataset is used to find the gradient by dividing the set into small subsets. These subsets are called mini batches and they are used to update the parameters in each iteration by taking one step. The entire training dataset is passed by using mini batches in one epoch. The training consists of many epochs. The mini batch size and the number of epochs should be determined before training. Sometimes, the gradient descent algorithm might oscillate around the local optima; therefore a momentum term is added to prevent this oscillation [24, 25]. The SGD update with momentum is as follows:

$$\theta_{k+1} = \theta_k - \alpha \nabla E(\theta_k) + \gamma(\theta_k - \theta_{k-1}), \quad (5)$$

where $\gamma$ is the impact of the previous gradient step in the current iteration.

L2 regularization term is added to the weights of the error function to reduce overfitting [24, 25]. The loss function with regularization is as follows:

$$E_R(\theta) = E(\theta) + \lambda \nabla \Omega(w), \quad (6)$$

where $w$ is the weight vector, $\lambda$ is the regularization coefficient, and $\Omega(w)$ is the regularization function.

$$\Omega(w) = \frac{1}{2} w w^T. \quad (7)$$

The error function is the cross-entropy function for 1-of-$k$ mutually exclusive classes as shown in the following equation:

$$E(\theta) = -\sum_{i=1}^{n} \sum_{j=1}^{k} t_{ij} \ln y_j(x_i, \theta), \quad (8)$$

where $\theta$ is a vector of parameters, $t_{ij}$ indicates that the $i$th sample is linked to the $j$th class, and $y_j(x_i, \theta)$ is the $i$th sample's output and can be formulated as a probability. The activation function of the output is the soft-max function:

$$y_r(x, \theta) = \frac{\exp(a_r(x, \theta))}{\sum_{j=1}^{k} \exp(a_j(x, \theta))}, \quad (9)$$

where $0 \le y_r \le 1$, $\sum_{j=1}^{k} y_j = 1$.

In this work, a fixed learning rate of 0.01 is used. The size of the mini batch is 300. The number of epochs is 30. The initial weights are produced by a Gaussian distribution with zero mean and standard deviation of 0.01. The initial bias value is zero. The momentum value is 0.9. The L2 regularization coefficient is 0.0001.

*(2) Pretrained CNN Model [26, 27].* "AlexNet" CNN with 5 convolutional layers has already been trained on ImageNet dataset. The network was trained for 90 cycles through the training set of 1.2 million images. This training took five to six days on two NVIDIA GTX 580 3 GB GPUs. After training, this model is used to extract discriminative features. The weights of the whole layers in the networks are fixed. The last fully connected layer, which is used to classify objects to 1000 classes, was removed. This pretrained model is able to extract 4096 features from each image. These features are used to train the various classifiers connected directly to the output of the seventh layer.

*2.2.2. Extreme Learning Machine [28].* Extreme learning machine (ELM) is a neural network that contains only one hidden layer. It possesses high generalization and highly efficient learning rate as its characteristics. These contribute to the success of this learning method. The biases and weights of the hidden layers are set randomly but the weights of the outputs are calculated analytically

$$f(x) = \sum_{i=1}^{L} F_i(x, W_i, b_i) \cdot \beta_i, \quad W_i \in R^d,\ b_i, \beta_i \in R, \quad (10)$$

where $F_i(\cdot)$ is an activation function of the $i$th hidden node, $W_i$ is an input weight, $b_i$ is a bias, and $\beta_i$ is the weight applied on the output. $L$ neurons are used in the hidden layer.

$$\beta = U^{\dagger} T,$$
$$\beta = U^T \left( \frac{1}{\lambda} + U \cdot U^T \right)^{-1} \cdot T, \quad (11)$$

where $U$ is an output of the hidden layer, $U^{\dagger}$ is the Moore–Penrose generalized inverse of a matrix, $T$ is a target, and $\lambda$ is a regulation coefficient.

*2.2.3. Hierarchical ELM for Feature Learning [13].* When dealing with visual data such as series of images, a deep architecture of extreme learning machine is required. Hierarchical extreme learning machine (HELM) is a recent deep model that is used to learn features automatically. It is utilized as a main block before the classifier to improve the system performance in terms of accuracy. This architecture can achieve self-taught feature learning via unsupervised ELM-based sparse encoder. HELM provides improved generalization and reduced learning time. The ELM-based sparse encoder is built using a fast iterative shrinkage thresholding algorithm (FISTA). This encoder is used as a basic component for HELM. Deep architecture is then achieved by stacking multiple encoders. It guarantees improved data recovery and reduces the testing time by reducing the number of neural nodes. Please refer to [13] for more details concerning HELM. HELM does not require the encoder's weights to be fine-tuned iteratively. This is the main reason why it is able to significantly reduce the time used for learning/training.

## 3. Experimental Results

*3.1. Dataset Description [29].* The University of Central Florida (UCF) published three types of datasets which are captured using an aerial camera, a Rooftop camera, and a Ground camera (ARG). Each dataset contains multiple human actions captured from multiple views. The datasets consist of ten actions performed by twelve persons. This paper focuses on the aerial camera dataset and five different activities: digging, waving, throwing, walking, and running.

This aerial dataset is considered as one of the most challenging datasets because the image samples are vary in activities, positions, orientations, viewpoints, cloth color, and scale. Besides that, the altitude of the camera varies according to the moving airborne platform. Figures 2, 3, and 4 show different samples of human and nonhuman objects in various situations.

In this paper, we focus on the usage of videos captured using an aerial camera. A high-definition camera (1920 × 1080 pixels at 60 fps) is mounted onto the payload platform of a helium balloon. For each activity, there are forty eight videos. Ten actions are performed four times by each person: boxing, carrying, clapping, digging, jogging, open-close trunk, running, throwing, walking, and waving. The actions are performed in various directions with three cars parked also in various directions.

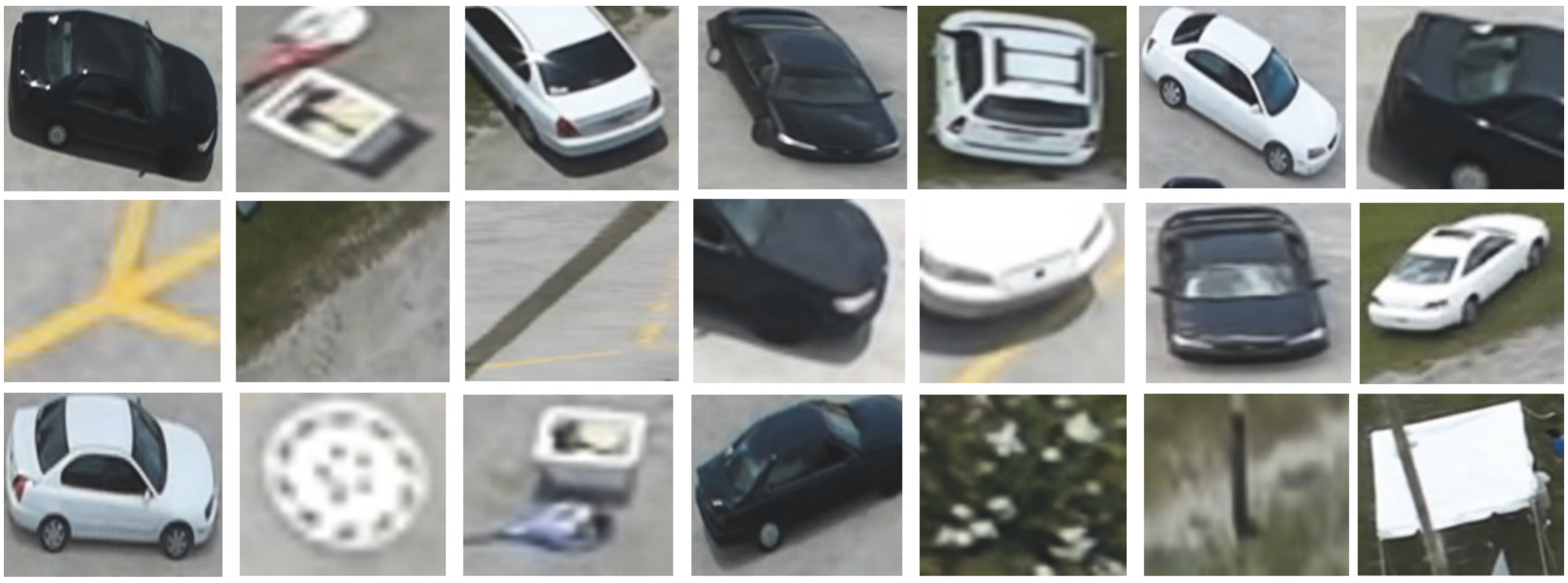

Figure 2: A few samples of nonhuman objects detected by the optical flow model. The images actually have different sizes but they are resized to the same size.

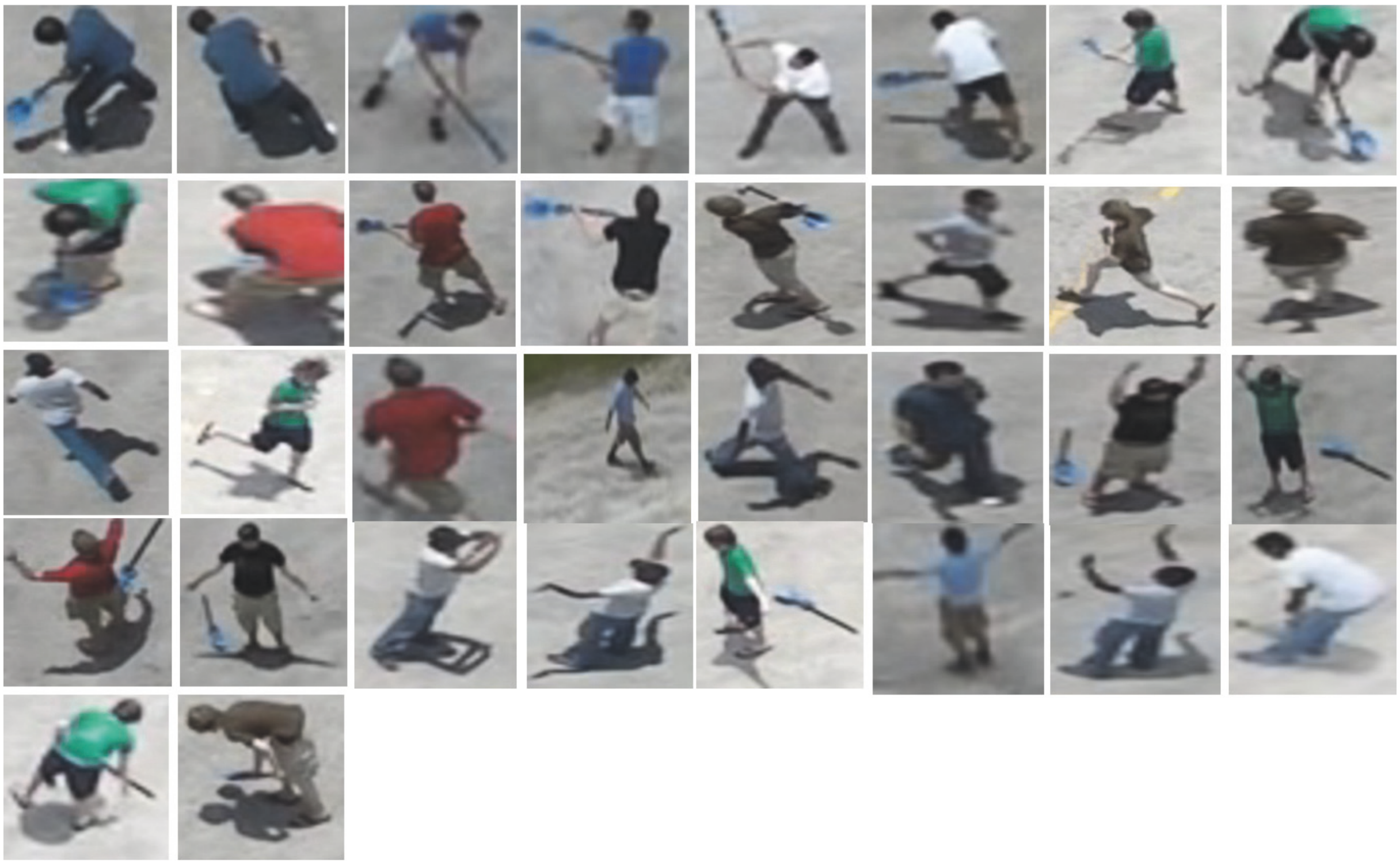

Figure 3: A few samples of human objects with different activities, positions, orientations, viewpoints, cloth color, and scale detected by the optical flow model. The images actually have different sizes but they are resized to the same size.

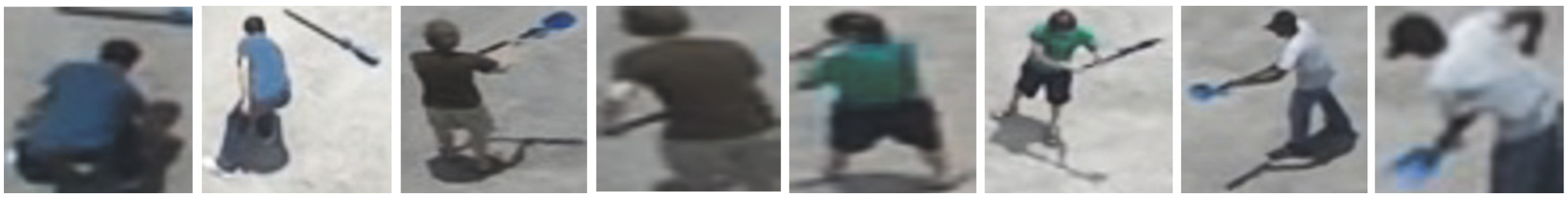

Figure 4: A few samples of human objects with varying altitudes due to aerial movement. The images actually have different sizes but they are resized to the same size.

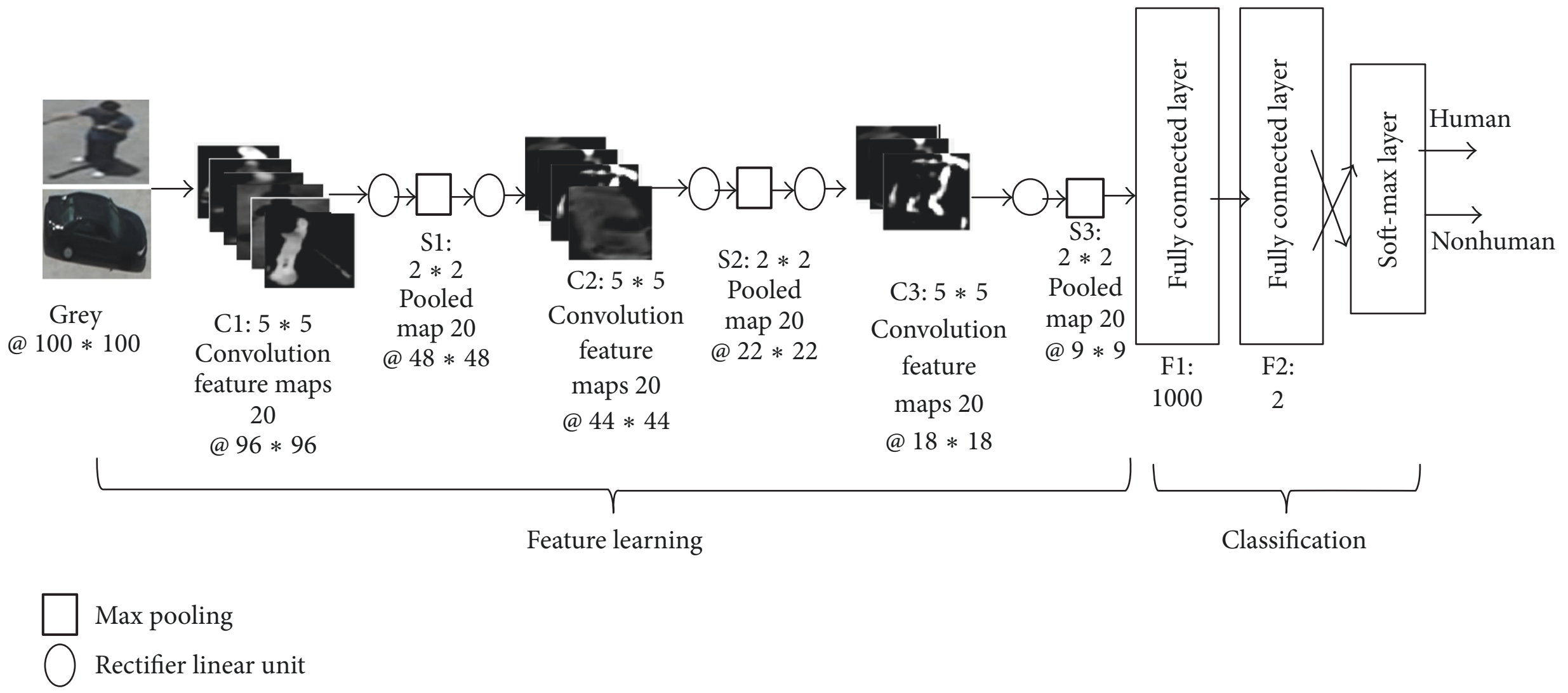

Figure 5: Structure of S-CNN model for human detection.

For experimental purposes, the performance of detection methods is evaluated for five different actions: digging, waving, throwing, walking, and running. This is sufficient as the main interest is to evaluate the performance of the proposed approach by using various deep models to classify aerial datasets which are highly challenging.

For this work, 48 (videos for each activity) ∗ 5 (activities) = 240 videos are used as training and testing data. 12 persons repeat the same activity 4 times. Eight of twelve persons are used for training and four persons are used for testing. The results are 160 videos for training and 80 videos for testing.

To reduce the number of samples used for training, not all frames in the video are used but only one frame from every ten frames is taken for training. These frames are used after applying optical flow to find moving patches within an image. One of these patches contains a human. All patches in one frame are used for training. Because the size of patches is not equal as a result of varying altitudes when airborne, these patches are resized to 227 ∗ 227 pixels in pretrained CNN and 100 ∗ 100 pixels in S-CNN and HELM before being processed by the deep models to extract the discriminative features.

The problem that appears in our training and testing samples is that the number of positive (human) and negative (nonhuman) samples is unbalanced. This is mainly caused by the optical flow stage that detects human and other regions in the images which do not have any human. To further clarify this issue, in each frame that is processed, only one patch is extracted for humans (via optical flow) and multiple patches are extracted for the others (i.e., nonhumans). The total number of samples is 26541. It includes 5862 positive samples and 20679 negative samples. However, the most interesting fact is that this problem does not affect our results. Even with an unbalanced dataset (between positive and negative samples), the proposed systems are still robust and are able to produce a high classification accuracy.

The experiment was implemented in Matlab2016a on a desktop computer. For HELM, the Intel core i7 @ 3.5 GHz CPU was used. For supervised CNN and pretrained CNN, the NVIDIA GetForce GTX 950 GPU was used. Both experiments were conducted in the Windows 8.1 (64 bits) environment.

*3.2. Supervised CNN Implementation.* The optimal CNN structure is shown in Figure 5. The architecture consists of 17 layers including an input layer, three convolutional layers, three max pooling layers, six rectifier linear unit layers, and finally two fully connected layers with a soft-max layer. The input layer contains a grey image with 100 ∗ 100 pixels. Each convolutional layer has twenty feature maps with 5 ∗ 5 convolution filters. The first convolutional feature map is 96 ∗ 96. The second is 44 ∗ 44. Then, the third one is 18 ∗ 18. The layers of max pooling have 2 ∗ 2. The first max pooling produces 48 ∗ 48 feature maps. The second one gives 22 ∗ 22 features maps. Then, the third one has 9 ∗ 9 feature maps. The six rectifier linear unit layers (Relu) are used between layers to clear the negative values. The last layers are fully connected with 1000 nodes. 2 nodes are connected to the soft-max layer to give two classes. Figure 6 shows the snapshots of feature maps after the first convolution layer C1, the second convolution layer C2, and the third convolution layer C3. Stochastic gradient descent was used to train the model with a mini batch size of 300. The learned features are extracted from the eighth layer "fc8" which is connected directly to a soft-max layer or SVM classifier to produce two classes: human and nonhuman.

The supervised CNN layers are as follows:

(1) Image input layer with 100 ∗ 100 pixels grey image.

(2) Convolution layer: 20 feature maps of 5 ∗ 5.

(3) Relu layer.

(4) Max pooling layer: pooling regions of size [2, 2] and returning the maximum of the four.

(5) Relu layer.

(6) Convolution layer: 20 feature maps of 5 ∗ 5.

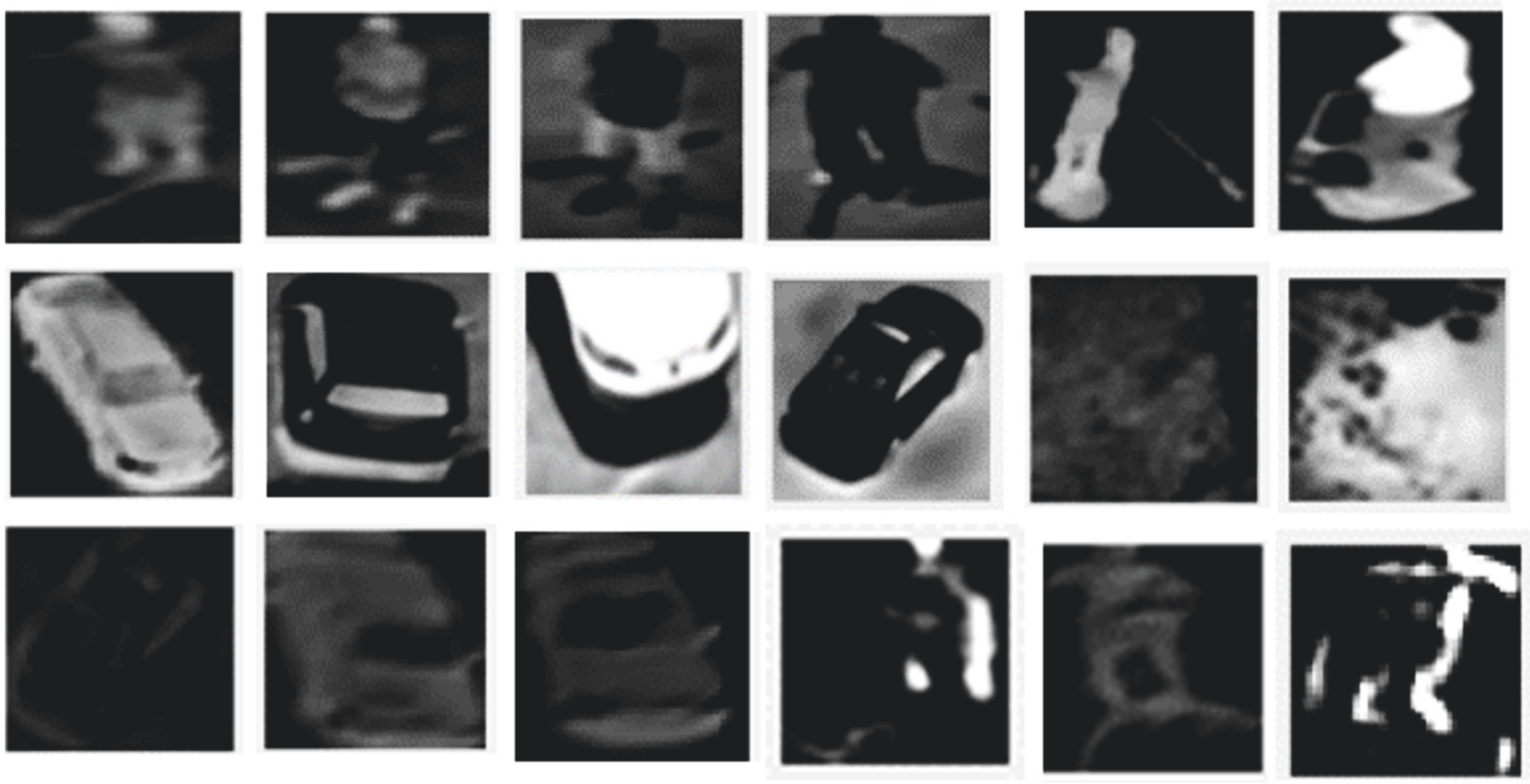

Figure 6: Snapshots of feature maps: the upper two rows show feature maps after the first convolution layer C1 20 ∗ (5 ∗ 5); the lowest row show feature maps after the second convolution layer C2 20 ∗ (5 ∗ 5).

(7) Relu layer.

(8) Max pooling layer: pooling regions of size [2, 2] and returning the maximum of the four.

(9) Relu layer.

(10) Convolution layer: 20 feature maps of 5 ∗ 5.

(11) Relu layer.

(12) Max pooling layer: pooling regions of size [2, 2] and returning the maximum of the four.

(13) Fully connected layer: with 1000 nodes.

(14) Relu layer.

(15) Fully connected layer with 2 classes.

(16) Soft-max layer.

(17) Classification layer.

*3.3. Pretrained CNN Implementation.* MatConvNet, Convolutional Neural Networks for MATLAB, is a toolbox that was designed for simple, flexible, and easy use of CNN building blocks. It provides MATLAB functions for computing linear convolutions with filter banks, feature pooling, and many more. By using MatConvNet, new CNN architectures can be built easily. "AlexNet" is one of those models which can be downloaded from MatConvNet [27]. A CUDA-capable GPU card is required to run this model. Figure 7 shows the architecture of the AlexNet model. The input color RGB image is resampled to 227 ∗ 227 ∗ 3 pixels. The networks include five convolutional layers, ReLU layers, max pooling layers, three fully connected layers, a soft-max layer, and a classification layer. The last fully connected layer is removed. The features are extracted from the seventh layer "fc7" which is connected directly to an SVM classifier to produce two classes: human and nonhuman. The activation function, which is used to extract features, is computed on the GPU. Stochastic gradient descent was used to train the model with a mini batch size of 32 to ensure that the CNN and image data fit into GPU memory. The number of extracted features is 4096. A multiclass SVM classifier is used to classify the features into human and nonhuman classes.

*3.4. HELM Implementation.* The optimal HELM structure is shown in Figure 8. The architecture consists of three modules including two sparse ELM-based autoencoders, and an ELM-based classifier. The input contains a grey image with 100∗100 pixels. In the first sparse autoencoder, there are 1000 neurons. The second sparse autoencoder also has 1000 neurons. The last module, which is a classifier, consists of 12,000 nodes in the hidden layers that are connected to 2 nodes in the output to produce two classes.

*3.5. Accuracy Analysis.* Table 1 shows the accuracy of each of the ten testing configurations. In each configuration, four persons are used in the testing data after applying the four-person-out cross-validation approach. The ten testing configurations are as follows:

Testing configurations: {p1, p2, p3, p4} where p1, p2, p3, and p4 are 4 testing persons.

{1, 2, 3, 4} {5, 6, 7, 8} {9, 10, 11, 12} {1, 3, 5, 7} {2, 4, 6, 8} {1, 4, 7, 10} {2, 5, 8, 11} {3, 6, 9, 12} {1, 5, 9, 12} {1, 6, 11, 12}.

The table compares the average accuracies of the proposed deep models in terms of accuracy. Pretrained CNN was found to outperform S-CNN and HELM with an average accuracy of 98.09%. Supervised CNN produces an average accuracy of 95.6% with the soft-max classifier and 91.7% with the support vector machine (SVM) classifier. HELM produces an average accuracy of 95.9%.

The leave four-out cross-validation model was used for average accuracy calculation. The experiment is repeated 10

Table 1: Accuracy comparison between CNN_soft-max, CNN_SVM, pretrained CNN, and HELM.

| Cross validation (4 persons out of 12) | Supervised CNN + soft-max | Supervised CNN + SVM | Pretrained CNN + SVM | HELM |
|---|---|---|---|---|
| 1 | 94.0542 | 91.8966 | 97.7170 | 94.7065 |
| 2 | 97.3726 | 91.9131 | 97.6115 | 97.9186 |
| 3 | 96.1542 | 92.9835 | 98.5988 | 96.0008 |
| 4 | 97.1056 | 92.8199 | 97.8509 | 97.4410 |
| 5 | 94.6982 | 91.1981 | 97.9344 | 94.9851 |
| 6 | 95.7974 | 91.2816 | 98.4077 | 95.7583 |
| 7 | 91.7131 | 87.8823 | 98.0749 | 92.3971 |
| 8 | 96.5649 | 92.7828 | 98.5311 | 96.5765 |
| 9 | 97.0207 | 93.6128 | 98.0556 | 97.3239 |
| 10 | 95.5184 | 90.6524 | 98.1257 | 95.7467 |
| Average accuracy | 95.5999% | 91.7023% | 98.0908% | 95.8855% |

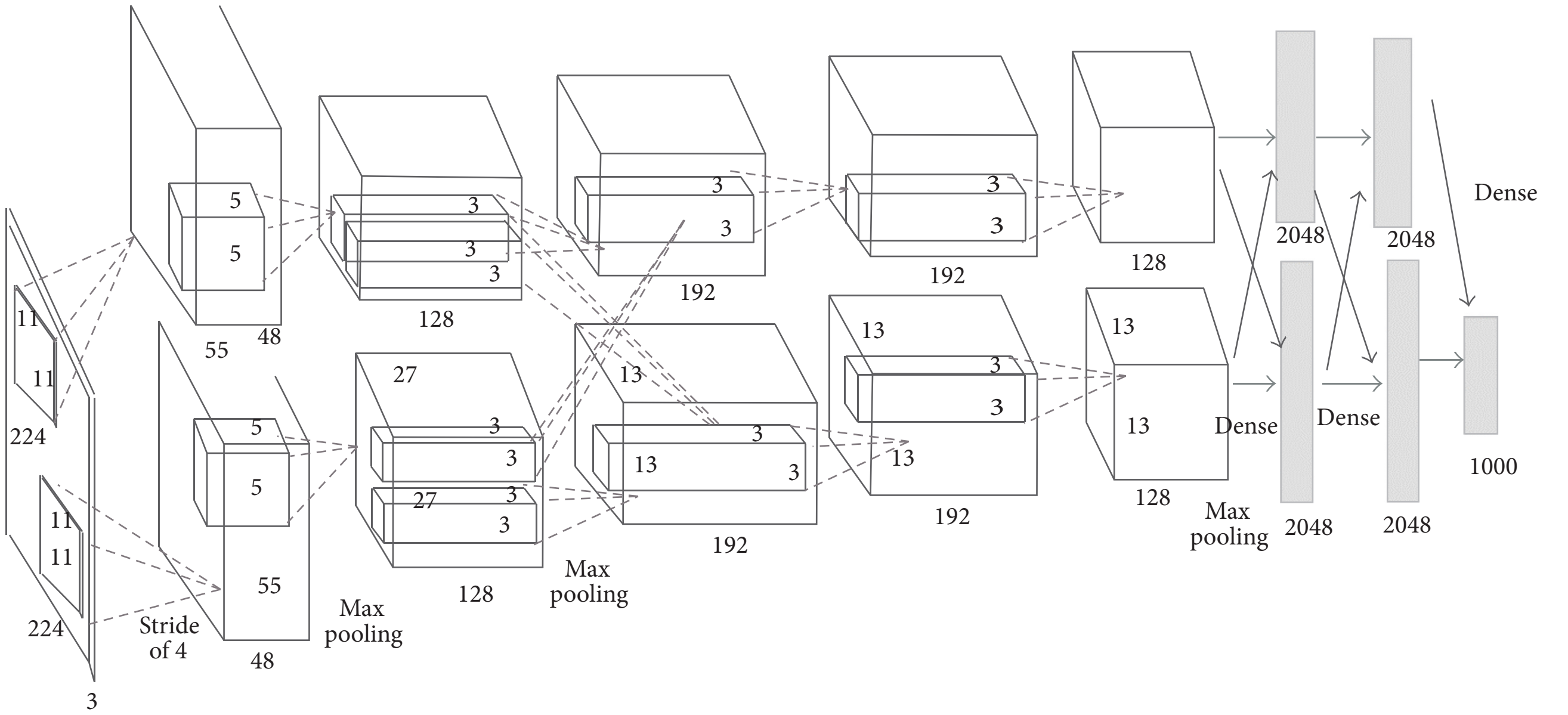

Figure 7: The architecture of the pretrained CNN model based on AlexNet [12].

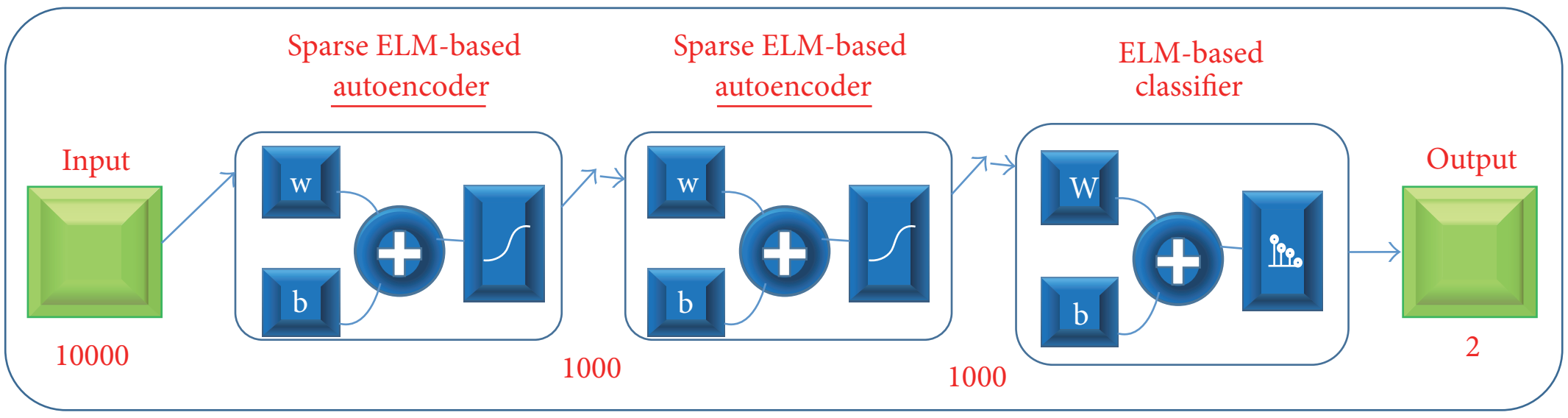

Figure 8: Deep model based on hierarchical ELM for human detection.

times. The average value from these experiments is taken as the final result.

Each time and for each activity, 32 videos are used as training data, and 16 videos are used as testing data. The testing video consists of a four new persons who perform the same activity 4 times. These four persons do not appear in the training data. The process is repeated for ten different configurations including four new persons out of twelve persons, then the average is calculated. Table 2 presents the number of training and testing samples in the UCF-ARG aerial dataset.

The confusion matrices for the last configuration $\{1, 6, 11, 12\}$ as testing data are shown in Figure 9. The matrices display the testing accuracy for four persons out of twelve. Each person repeats the five actions four times. So the number of testing videos is 80 out of 240.

Table 2: The number of training and testing data in the UCF-ARG aerial dataset.

| Cross validation (4 persons out of 12) | Number of training data | Number of testing data |
|---|---|---|
| 1 | 18569 | 7972 |
| 2 | 17749 | 8792 |
| 3 | 16764 | 9777 |
| 4 | 18071 | 8050 |
| 5 | 17374 | 8714 |
| 6 | 18006 | 7662 |
| 7 | 15360 | 10233 |
| 8 | 16733 | 8646 |
| 9 | 15607 | 9566 |
| 10 | 16183 | 8323 |

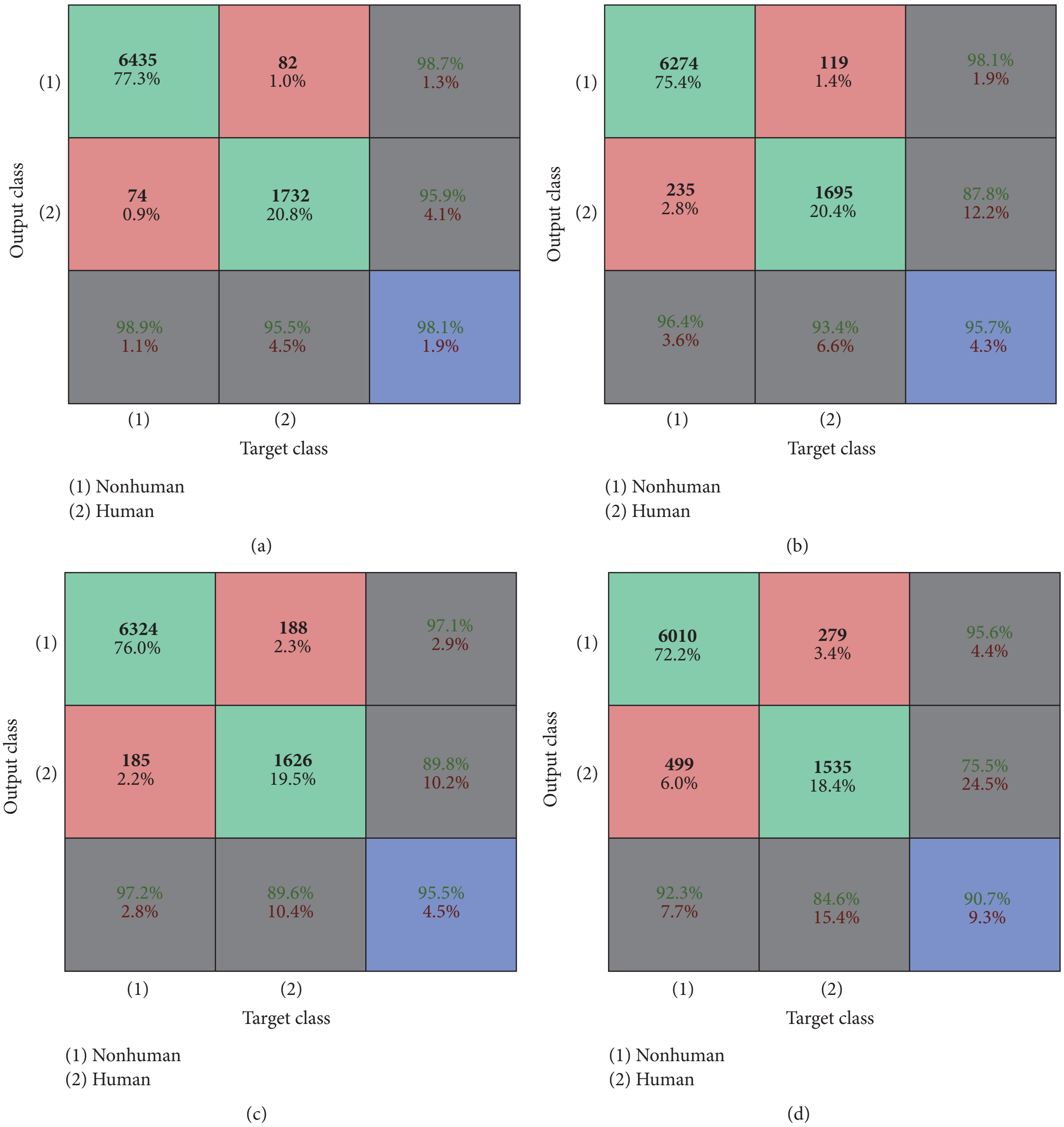

Figure 9: (a) Testing accuracy for pretrained CNN, (b) testing accuracy for HELM, (c) testing accuracy for supervised CNN + soft-max, and (d) testing accuracy for supervised CNN + SVM.

Table 3: The training and testing speed comparison between supervised CNN and HELM.

| Methods | Training time [s] | Testing time for one sample [s] |
|---|---|---|
| S-CNN (GPU) | 770 | 0.045 |
| HELM (CPU) | 445 | 0.1 |

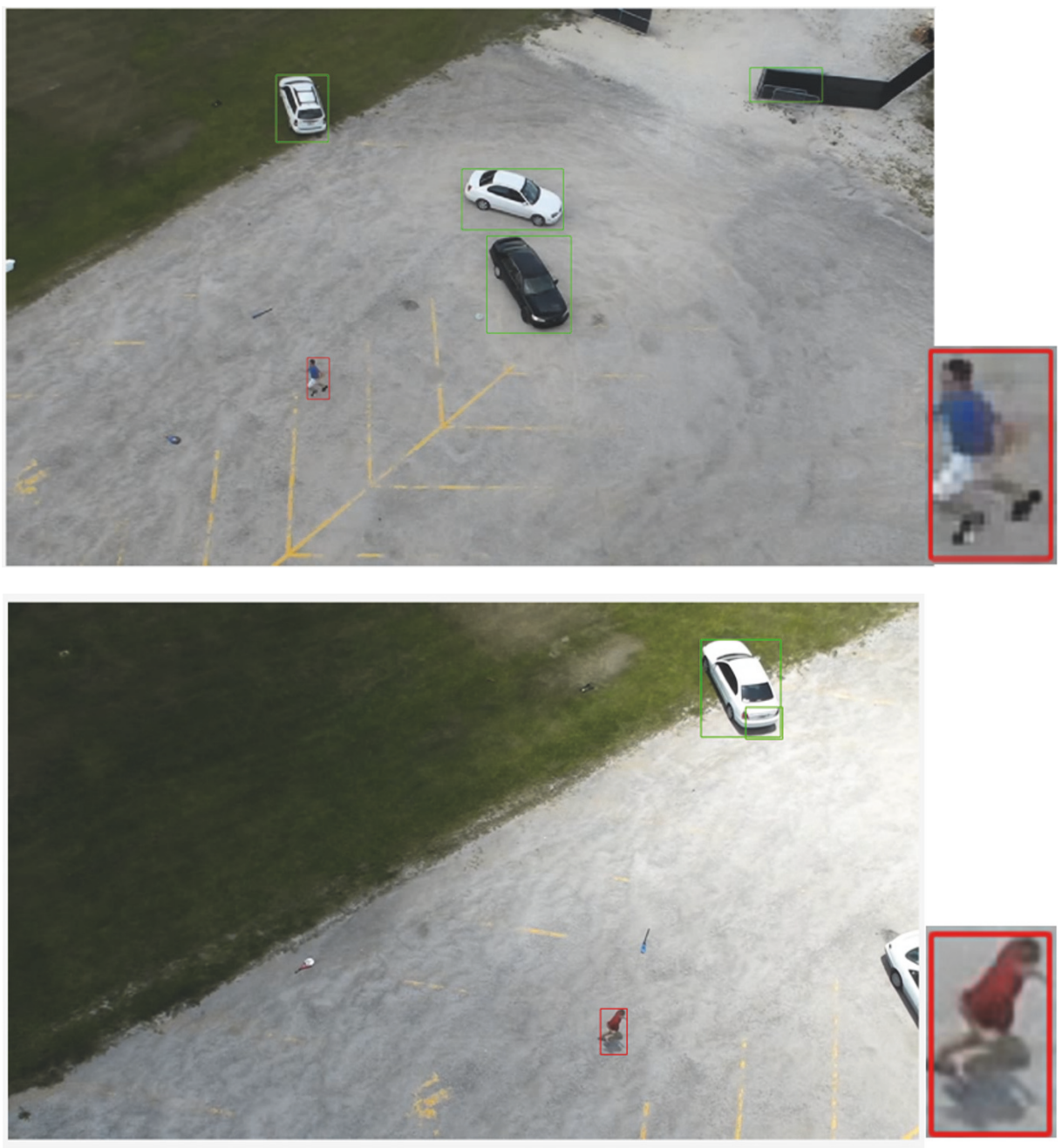

Figure 10: Humans running in the UCF-ARG dataset: the green boundary boxes in the image are the results after using optical flow. The red boundary boxes are the results after applying human classification based on deep models.

*3.6. Speed Analysis.* Table 3 shows the difference between two feature learners (HELM and supervised CNN) in terms of training and testing time. The proposed HELM-based detector outperforms S-CNN in terms of training speed even though HELM uses CPU whereas S-CNN uses GPU. This results from utilizing hierarchical extreme learning machine as a fast deep model that does not require fine tuning of weights iteratively. This advantage of HELM gives a chance to be implemented in real time on low cost embedded system.

Figures 10, 11, and 12 show the original frames with green and red boundary boxes. The green boundary boxes result from the optical flow model. Different objects are detected by optical flow and surrounded by green boxes. The red boundary box is drawn after using a deep model to detect only humans. The simulation videos have been uploaded to the YouTube website. Please refer to the following link:

https://www.youtube.com/channel/UCvEvheiIvcV_n_NvK3l4vQA.

The model is trained for five activities (digging, waving, throwing, walking, and running). In Figure 13, multiple human detections are tested. This generalization performance is suitable for detecting persons which are performing different untrained activities such as clapping, boxing, carrying, and jogging. Figure 14 shows multiple persons in one frame. Each person performs a different activity. The proposed models are able to detect all humans in the frame and draw boundary boxes around them.

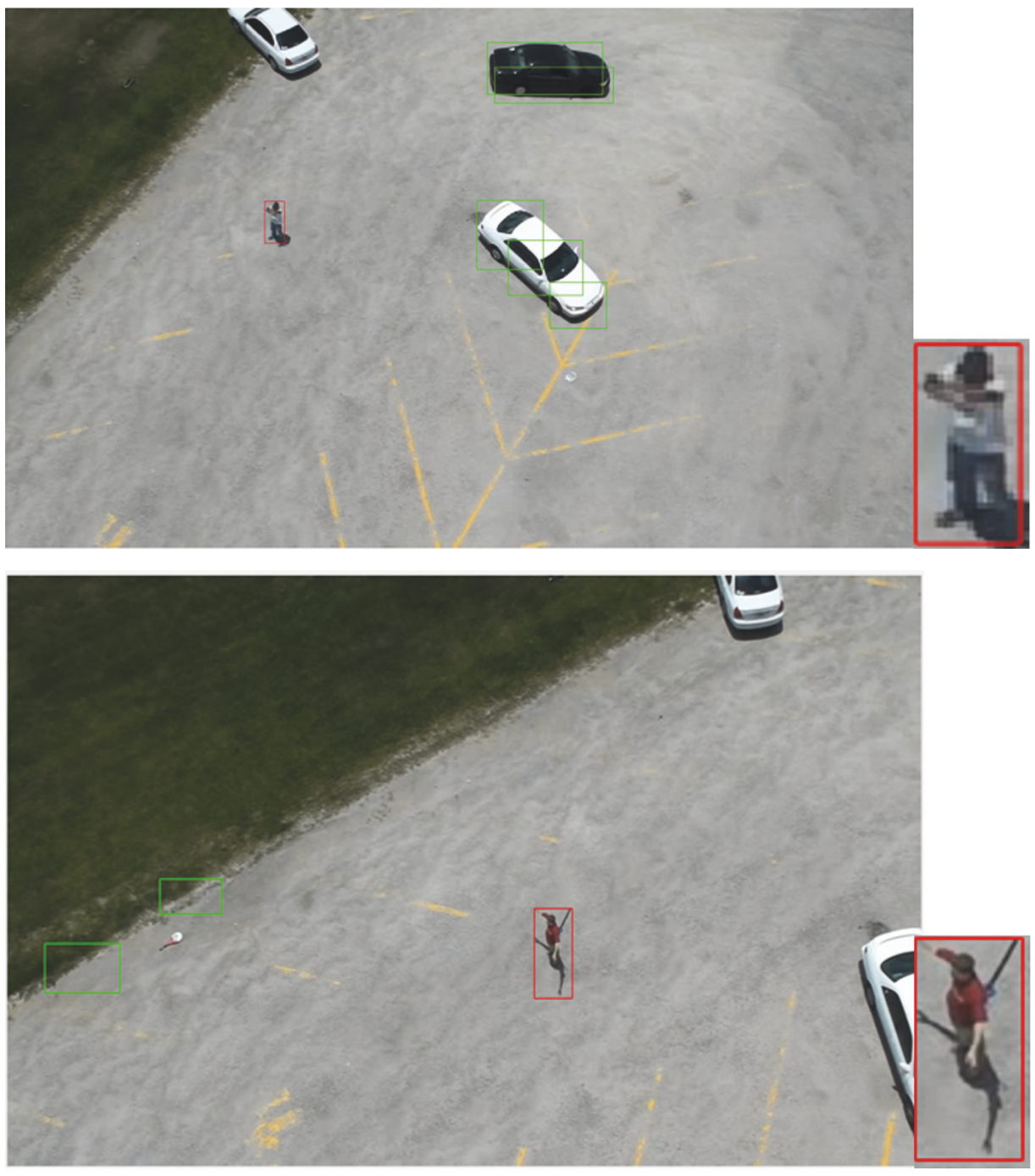

Figure 11: Humans waving in the UCF-ARG dataset: the green boundary boxes in each image are the results after using optical flow. The red boundary boxes are the results after applying human classification based on deep models.

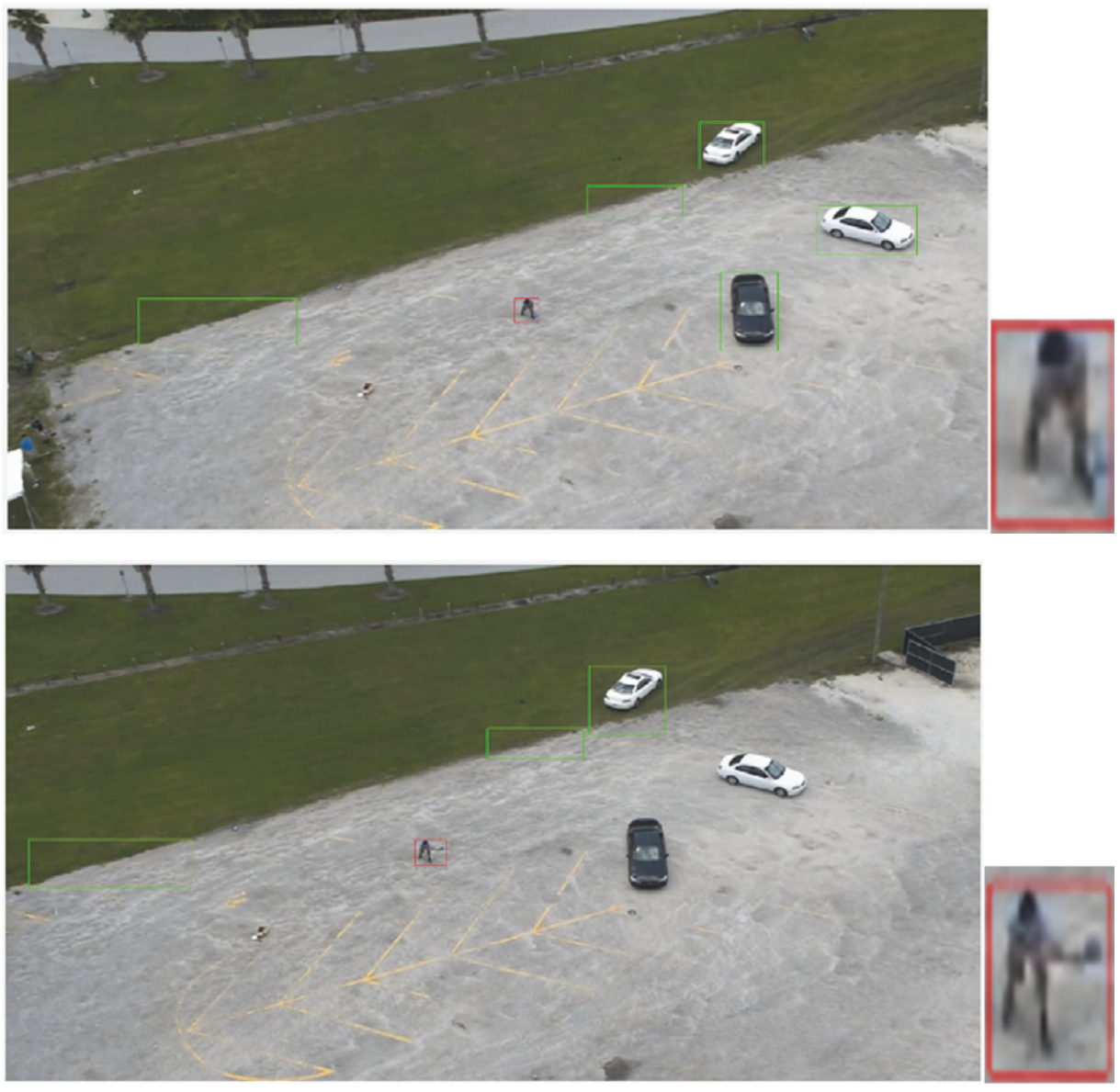

Figure 12: Humans digging in the UCF-ARG dataset: the green boundary boxes in each image are the results after using optical flow. The red boundary boxes are the results after applying human classification based on deep models.

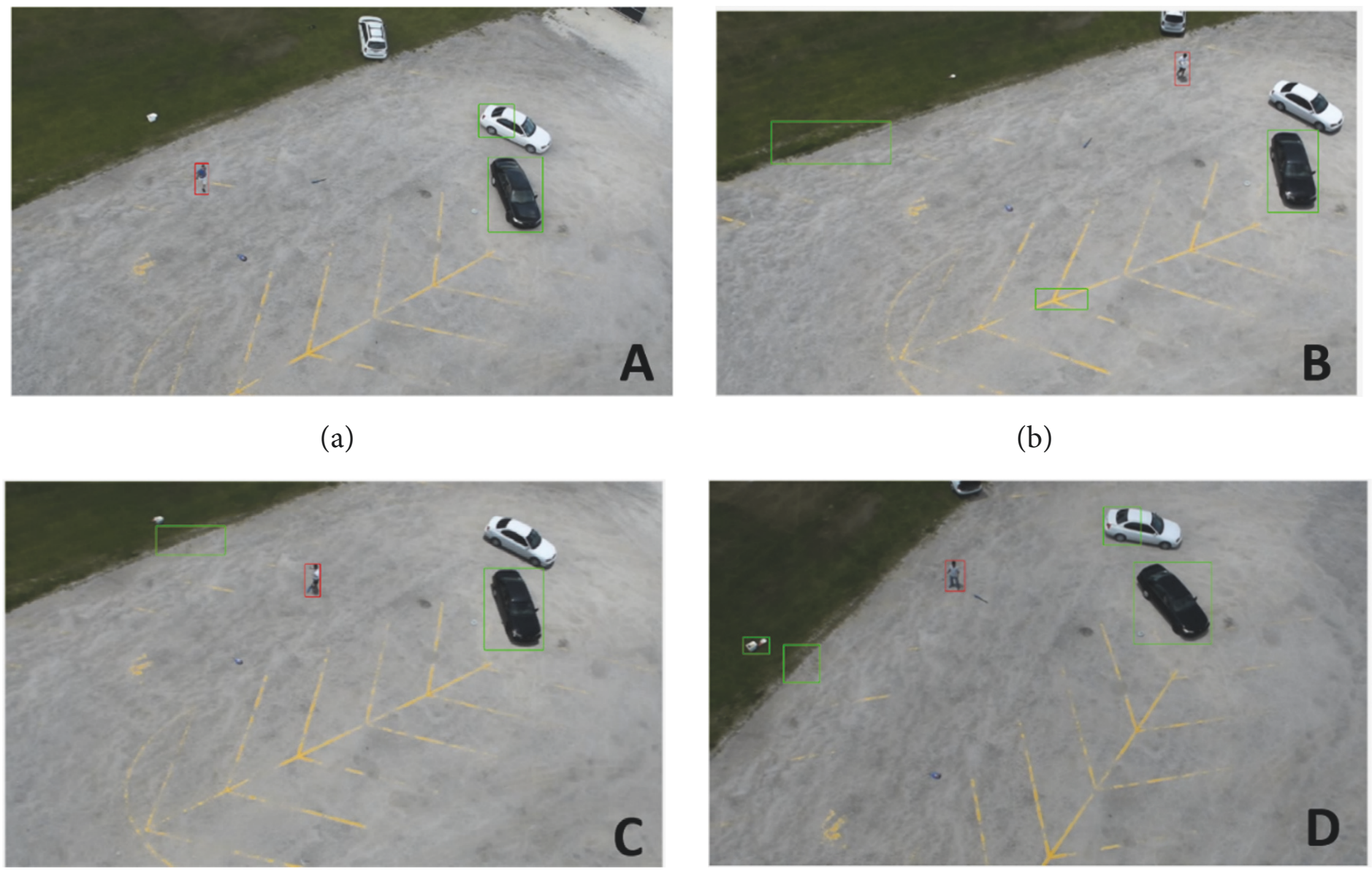

(a) (b) (c) (d)

Figure 13: Detecting humans with different untrained actions from the UCF-ARG dataset ((a) for jogging. (b) for carrying. (c) for boxing. (d) for clapping). Green boundary boxes in each image are the results after using optical flow. The red boundary boxes are the results after applying human classification based on deep models.

## 4. Discussion and Conclusion

In this paper, the implementation of three deep models for human detection was demonstrated. The performance was evaluated for five human actions (digging, waving, throwing, walking, and running). The models were trained and tested on the public UCF-ARG aerial dataset. The challenging part of this dataset is the size of human patches which vary according to the altitude of the moving airborne platform and the multiple viewpoints of humans in the same video. Also, due to the usage of optical flow, the training and testing data was unbalanced in terms of the number of positive and negative samples.

The results of this work can be summarized as follows:

(1) The quality of the stabilization method (optical flow) is important in our proposed systems as a first stage before applying the deep models for human, nonhuman classification.

(2) The proposed systems solve the trade-off problem between high accuracy and speed of detection. The pretrained deep model was found to outperform S-CNN and HELM in terms of accuracy with an average accuracy of 98.09%. The pretrained CNN and S-CNN models were implemented on a GPU. HELM was more time efficient than S-CNN because it does not require iterative fine tuning of the weights. This is because the weights are generated randomly. Due to this, the training time is reduced to 445 s with a normal CPU. At the same time, it produces a good average classification accuracy of 95.9%. HELM is recommended for use in embedded systems that require high accuracy with low cost.

The advantages of the proposed systems are as follows:

(i) The proposed system can detect humans automatically and does not need a manual detection threshold to select one that has the highest true positive rate.

(ii) The generalization of both deep models is able to detect humans accurately in all 80 videos that are not in the training data.

(iii) The proposed system achieves real-time performance for testing live-captured videos because the optical flow model only utilizes two successive frames to find motion. Moreover, deep models only require a single frame to classify the optical flow patches as human or nonhuman (i.e., human detection).

(iv) The proposed deep models are robust against various activities, positions, orientations, viewpoints, cloth color, scale, and altitudes.

The drawback of the proposed system is that it is highly dependent on the quality of the optical flow processing stage. Adding tracking to the whole pipeline for human detection can reduce this dependency and improve the overall accuracy. For a future work, we will integrate tracking that makes use of initially extracted training regions around humans as positive samples and other regions as negative samples. As the result is highly accurate and efficient, we will utilize the results of human detection demonstrated in this paper for human action recognition to map each activity with a specific action class.

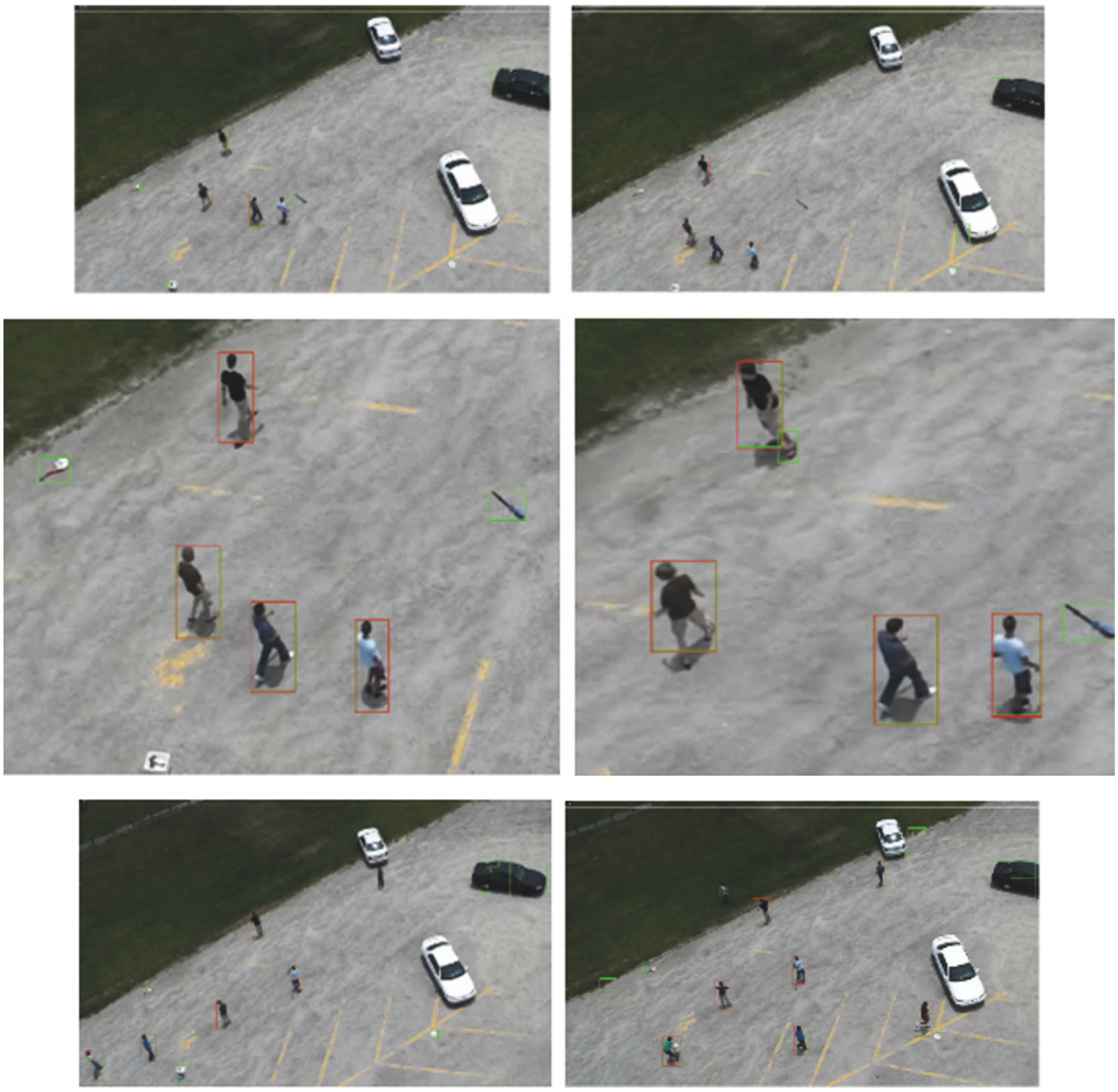

Figure 14: Human detection for multiple human with different actions from UCF-ARG dataset: Green boundary boxes in each image are the results after using optical flow. The red boundary boxes are the results after applying human classification based on deep models. The first and third rows are the original frames. The second row is the first row after zooming in on the persons.

## Conflicts of Interest

The authors declare that they have no conflicts of interest.

## Acknowledgments

This work is funded by the University of Malaya's Research Grant (UMRG), Grant no. RP030A-14AET, and the Fundamental Research Grant (FRGS), Grant no. FP061-2014A.

## References

[1] X. Wang, T. X. Han, and S. Yan, "An HOG-LBP Human Detector with Partial Occlusion Handling," in *Proceedings of the IEEE 12th International Conference on Computer Vision ICCV*, pp. 32–39, 2009.

[2] D. G. Lowe, "Distinctive image features from scale-invariant keypoints," *International Journal of Computer Vision*, vol. 60, no. 2, pp. 91–110, 2004.

[3] N. Dalal and B. Triggs, "Histograms of oriented gradients for human detection," in *Proceedings of the IEEE Computer Society Conference on Computer Vision and Pattern Recognition (CVPR '05)*, vol. 1, pp. 886–893, June 2005.

[4] P. Dollar, S. Belongie, and A. P. Perona, "The Fastest Pedestrian Detector in the West," in *Proceedings of the British Machine Vision Conference*, pp. 68.1-68.11, BMVA Press, 2010.

[5] Y. Lecun, L. Bottou, Y. Bengio, and P. Haffner, "Gradient-based learning applied to document recognition," in *Proceedings of the IEEE*, vol. 86, Issue: 11, pp. 2278–2324, 2002.

[6] A. Rikhtegar, M. Pooyan, and M. T. Manzuri-Shalmani, "Genetic algorithm-optimised structure of convolutional neural network for face recognition applications," *IET Computer Vision*, vol. 10, no. 6, pp. 559–566, 2016.

[7] N. Kalchbrenner, E. Grefenstette, and P. Blunsom, "A convolutional neural network for modelling sentences," in *Proceedings*

*of the 52nd Annual Meeting of the Association for Computational Linguistics, ACL 2014*, pp. 655–665, June 2014.

[8] Y. Kim, "Convolutional neural networks for sentence classification," in *Proceedings of the 2014 Conference on Empirical Methods in Natural Language Processing, EMNLP 2014*, pp. 1746–1751, qat, October 2014.

[9] R. Collobert and J. Weston, "A unified architecture for natural language processing: deep neural networks with multitask learning," in *Proceedings of the 25th International Conference on Machine Learning*, pp. 160–167, ACM, July 2008.

[10] A. Casamitjana, S. Puch, A. Aduriz, and V. Vilaplana, "3D Convolutional Neural Networks for Brain Tumor Segmentation," in *Proceedings of the MICCAI Challenge on Multimodal Brain Tumor Image*, 2016.

[11] M. Baccouche, F. Mamalet, C. Wolf, C. Garcia, and A. Baskurt, "Sequential Deep Learning for Human Action Recognition," in *In Human Behavior Understanding*, pp. 29–39, Springer, 2011.

[12] A. Krizhevsky, I. Sutskever, and G. E. Hinton, "Imagenet classification with deep convolutional neural networks," in *Proceedings of the 26th Annual Conference on Neural Information Processing Systems (NIPS '12)*, pp. 1097–1105, Lake Tahoe, Nev, USA, December 2012.

[13] J. Tang, C. Deng, and G. B. Huang, "Extreme learning machine for multilayer perceptron," *IEEE Transactions on Neural Networks & Learning Systems*, vol. 27, no. 4, pp. 809–821, 2016.

[14] A. Angelova, A. Krizhevsky, V. Vanhoucke, A. Ogale, and D. Ferguson, "Real-Time Pedestrian Detection With Deep Network Cascades," in *Proceedings of British Machine Vision Conference (BMVC)*, 2015.

[15] H. Fukui, T. Yamashita, Y. Yamauchi, H. Fujiyoshi, and H. Murase, "Pedestrian detection based on deep convolutional neural network with ensemble inference network," in *Proceedings of the IEEE Intelligent Vehicles Symposium, IV 2015*, pp. 223–228, kor, July 2015.

[16] I. Cohen and G. Medioni, "Detecting and tracking moving objects for video surveillance," in *Proceedings of the 1999 IEEE Computer Society Conference on Computer Vision and Pattern Recognition (CVPR'99)*, pp. 319–325, June 1999.

[17] F. Schmidt and S. Hinz, "A Scheme for the Detection and Tracking of People Tuned for Aerial Image Sequences," in *Photogrammetric Image Analysis of the book series Lecture Notes in Computer Science (LNCS)*, vol. 6952 of *270*, pp. 270-257, 2011.

[18] A. W. Eekeren, J. Dijk, and G. Burghouts, "Detection and tracking of humans from an airborne platform," in *Proc. SPIE 9249, Electro-Optical and Infrared Systems: Technology and Applications*, vol. 9249, 2014.

[19] A. Fernández-Caballeroa, J. C. Castillo, J. M. Cantos, and R. M. Tomás, "Optical flow or image subtraction in human detection from infrared camera on mobile robot," *Robotics and Autonomous Systems*, vol. 58, no. 12, pp. 1273–1281, 2010.

[20] H. Han and M. Tong, "Human detection based on optical flow and spare geometric flow," in *Proceedings of the 2013 7th International Conference on Image and Graphics, ICIG 2013*, pp. 459–464, China, July 2013.

[21] http://spie.org/x41092.xml?ArticleID=x41092.

[22] J. L. Barron, D. J. Fleet, and S. S. Beauchemin, "Performance of optical flow techniques," *International Journal of Computer Vision*, vol. 12, no. 1, pp. 43–77.

[23] B. K. P. Horn and B. G. Schunck, "Determining optical flow," *Artificial Intelligence*, vol. 17, no. 1-3, pp. 185–203, 1981.

[24] C. M. Bishop, *Pattern Recognition and Machine Learning*, Springer, New York, NY, USA, 2006.

[25] K. P. Murphy, *Machine Learning: A Probabilistic Perspective*, The MIT Press, Cambridge, Mass, USA, 2012.

[26] J. Deng, W. Dong, and R. Socher, "ImageNet: a large-scale hierarchical image database," in *Proceedings of the 2009 IEEE Conference on Computer Vision and Pattern Recognition (CVPR)*, pp. 248–255, Miami, Fla, USA, June 2009.

[27] A. Vedaldi and K. Lenc, "MatConvNet-convolutional neural networks for MATLAB," arXiv preprint arXiv:1412.4564.

[28] G. B. Huang, Q. Y. Zhu, and C. K. Siew, "Extreme learning machine: theory and applications," *Neurocomputing*, vol. 70, no. 1–3, pp. 489–501, 2006.

[29] K. Reddy, UCF-ARG Data Set (http://crcv.ucf.edu/data/UCF-ARG.php).